\documentclass[10pt,twocolumn,letterpaper]{article}

\usepackage{cvpr}
\usepackage{times}
\usepackage{epsfig}
\usepackage{graphicx}
\usepackage{amsmath}
\usepackage{amssymb}
\usepackage{tabulary}
\usepackage{rotating}
\usepackage{multirow}
\usepackage{subcaption}
\usepackage{booktabs}
\usepackage{chngcntr}
\usepackage[toc,page]{appendix}
\renewcommand\appendixtocname{Appendix}
\renewcommand\appendixpagename{Appendix}

\DeclareMathOperator*{\argmin}{arg\,min}
\newcommand{\ignore}[1]{}
\usepackage{multirow}
\usepackage{makecell}%
\setcellgapes{3pt}%

\newcommand{\deflen}[2]{%
    \expandafter\newlength\csname #1\endcsname
    \expandafter\setlength\csname #1\endcsname{#2}%
}

\usepackage[pagebackref=true,breaklinks=true,letterpaper=true,colorlinks,bookmarks=false]{hyperref}

\cvprfinalcopy %

\def\cvprPaperID{25} %

\ifcvprfinal\fi
\begin{document}

\title{Role of Spatial Context in Adversarial Robustness for Object Detection}

\author{
  Aniruddha Saha\footnotemark[1] \qquad 
  Akshayvarun Subramanya\thanks{Equal contribution} \qquad  Koninika Patil \qquad Hamed Pirsiavash\\
  University of Maryland, Baltimore County \\
  \texttt{\{anisaha1, akshayv1, koni1, hpirsiav\}@umbc.edu} \\
}

\maketitle

\begin{abstract}
The benefits of utilizing spatial context in fast object detection algorithms have been studied extensively. Detectors increase inference speed by doing a single forward pass per image which means they implicitly use contextual reasoning for their predictions. However, one can show that an adversary can design adversarial patches which do not overlap with any objects of interest in the scene and exploit contextual reasoning to fool standard detectors. In this paper, we examine this problem and design category specific adversarial patches which make a widely used object detector like YOLO blind to an attacker chosen object category. We also show that limiting the use of spatial context during object detector training improves robustness to such adversaries. We believe the existence of context based adversarial attacks is concerning since the adversarial patch can affect predictions without being in vicinity of any objects of interest. Hence, defending against such attacks becomes challenging and we urge the research community to give attention to this vulnerability.
\end{abstract}

\section{Introduction}

\noindent The computer vision community has studied the role of spatial context in object detection for a long time. It is well-known that object detection performance is improved by considering information from the entire scene \cite{divvala2009empirical,torralba2003context,desai2011discriminative,barnea2019exploring}. Fast object detectors (YOLO \cite{redmon2016you}, SSD \cite{liu2016ssd}, and Faster-RCNN \cite{ren2015faster}) process the entire image in one forward pass to reduce inference time and hence implicitly use contextual information for their predictions. Some previously used object detectors (e.g., RCNN \cite{girshick2015fast}) do not use context since they do a forward pass for each region proposal separately. However, they are less accurate and have much slower inference compared to the fast detectors mentioned above.

In this paper, we study the scenario where an adversary can exploit the implicit contextual reasoning to fool object detectors in a practical setting. We show that category specific contextual adversarial patches (e.g. all categories in PASCAL VOC) can be designed which when pasted on the corner of an image, having no overlap with any objects of interest can make the detector blind to a specific object category chosen by the adversary, while not influencing the detections of objects of other categories. For example, if the category chosen is ``pedestrian'' in self-driving car applications, the attack may have a high impact on safety. Since our adversarial patch does not overlap with the object being attacked, we believe the attack works because the detector considers the influence of not only the object but also the surrounding objects and the scene. 

\begin{figure*}[!h]
\captionsetup{font=small}
  \begin{center}
  \begin{tabular}{| c c c c |}
\hline  
 \footnotesize{YOLOv2} &  \footnotesize \footnotesize{YOLOv2} & \footnotesize Grad-Defense & \footnotesize Grad-Defense \\
\footnotesize Detection & \footnotesize Adv patch Detection & \footnotesize{Detection} &  \footnotesize Adv patch Detection \\
\hline
\vspace{-.08in}
&&&\\
\vspace{-.05in}
\begin{sideways} \scriptsize ~~~~~~~~~~~~~~Target: Cat \end{sideways}
\includegraphics[width=.15\textwidth]{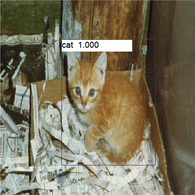}&

\includegraphics[width=.15\textwidth]{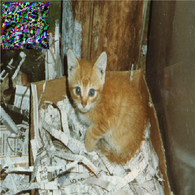}&
\includegraphics[width=.15\textwidth]{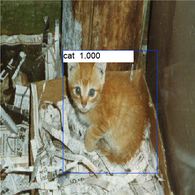}&
\includegraphics[width=.15\textwidth]{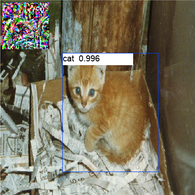}\\
& \small Cat fooled && \small Cat detected\\
&&&\\

\vspace{-.05in}
\begin{sideways} \scriptsize ~~~~~~~~~~~~~~Target: Bird \end{sideways}
\includegraphics[width=.15\textwidth]{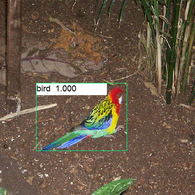}&

\includegraphics[width=.15\textwidth]{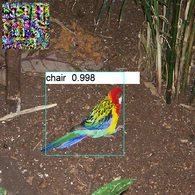}&
\includegraphics[width=.15\textwidth]{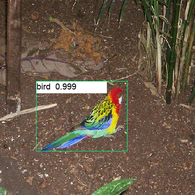}&
\includegraphics[width=.15\textwidth]{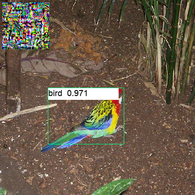}\\
& \small Bird fooled & & \small Bird detected\\
\hline
\end{tabular}
  \caption{\textbf{Per-image blindness attack and Grad-Defense results-} We compare the detection results of YOLOv2 with our Grad-Defense model. We attack each model separately. We see that our Grad-Defense model is less susceptible to context-based attack. The target category is written below each example. The patch is always on the top-left corner.}
\label{fig:perimagepatchfigure}
  \end{center}
   \vspace{-.35in}
\end{figure*}

We believe this is of paramount importance because:

\noindent (1) This poses a question to the research community with regards to improving accuracy versus robustness when using detectors that utilize context. The problem is challenging since there is no straightforward way of limiting fast state-of-the-art object detectors to avoid using context even at the cost of accuracy. As mentioned above, these detectors process the image only once for fast inference.
Also, in deep networks, it is difficult to limit the receptive field of the final layers so as to not cover the whole image.

\noindent (2) Adversarial patch attacks are easily reproducible in a practical setting as compared to standard adversarial attacks which add $\epsilon$-bounded perturbation to the whole image. One can simply print an adversarial patch and expose it to a self-driving car by pasting it on a wall, on a bumper, or even by hacking a road-side billboard. In contrast, regular adversarial examples need to perturb all pixel values. Though the perturbations are small, this is not very practical.

\noindent (3) Defense algorithms developed for regular adversarial examples\cite{madry2018towards,43405} are not necessarily suitable for adversarial patches since the patches are not norm-constrained, i.e., can have large perturbations in a limited number of dimensions. Hence, in the space of pixel values, the adversarial image, i.e., image + patch, can be very far from the original image. There are two standard defense frameworks for regular adversarial examples: (a) Training with adversarial examples: This is not suitable for our case since learning the patch is expensive. (b) Training with regularizers in the input space: This is not suitable either since such defense algorithms assume the noisy image is close to the original one in the pixel space. 

In this paper, we propose a defense which limits the use of context while training the object detector. We observe that it performs the best among all the other defense baselines.  We show an example in Figure \ref{fig:perimagepatchfigure} where our defense model is more robust to adversarial patch attack. We believe that this is a much needed step towards developing defense algorithms which fix the vulnerabilities introduced by adversarial patches. 

As an additional contribution, we investigate the efficacy of currently used explainability algorithms for object detectors. We observe that the well-known Grad-CAM algorithm \cite{selvaraju2016grad} is not suitable for localization tasks. We modify this algorithm to localize the explanation of each detection and use it to visualize the use of context before and after training using our defense algorithm. 

Our findings in this paper show that even though we believe in the richness of context in object detection, employing contextual reasoning can be a \emph{double-edged sword} that makes the object detection algorithms vulnerable to adversarial attacks. 

\section{Related Work}
\noindent \textbf{Vulnerability to Adversarial Attacks:}  Convolutional neural networks have been shown to be vulnerable to adversarial examples. Szegedy \etal \cite{intriguing-arxiv-2013} first discovered the existence of adversarial images. Gradient based techniques such as Fast Gradient Sign Method (FGSM) \cite{43405} and Projected Gradient Descent (PGD) \cite{madry2018towards} were used to create these adversarial examples. \cite{su2019one} showed that perturbing just one pixel is sufficient to fool classifiers. \cite{deepfool-cvpr-2016} presented an optimal way to create adversarial examples and later extended that to create an universal adversarial image \cite{moosavi2017universal}. However these attacks are not feasible for practical applications like IOT cameras and autonomous cars since we do not have access to all pixels and even perturbing single pixel that fools in multiple conditions becomes challenging. \cite{zajkac2018adversarial}  showed that we can fool classifiers  by adding an adversarial frame around the image that is trained to fool the classifier.  Apart from classification, object detection and semantic segmentation networks have also shown to be vulnerable to adversarial examples. \cite{xie2017adversarial,220580}  have shown that  adversarial examples can be created for object detectors as well. Fischer et al. \cite{Fischer2017AdversarialEF} showed the same for semantic segmentation.  

\noindent \textbf{Adversarial patches:} Adversarial patches were introduced in  \cite{brown2017adversarial,karmon2018lavan} which when pasted on an image can cause a classifier to output a target class. This poses a question as to whether object detection also has these vulnerabilities. Humans are rarely susceptible to mis-classifying objects in a scene if we introduce artifacts which do not overlap with the objects of interest. However, we show in this paper that object detection networks which use context can be fooled easily, taking advantage of this fact. In \cite{220580,thys2019fooling}, adversarial patches are pasted on top of the object (e.g. stop sign and person). This changes the appearance of the object and fools the detector. We consider a setting where the patch has no overlap with the object and thus highlight that contextual reasoning can lead to adversarial vulnerability making the threat model more challenging.
\cite{DBLP:journals/corr/abs-1806-02299} create a digital attack on object detectors where the pixel values are unconstrained. \cite{lee2019physical} designs a patch for YOLO which when presented in a scene suppresses all detections. But we show that category specific contextual adversarial patches can be designed which can make the detector blind to a specific object category chosen by the adversary and not strongly influence detections of objects of other categories. \cite{huangadversarial} create adversarial signboards which when placed below each instance of a stop sign fools the detector. Our patch can be far from all objects and can produce blindness for all the attacker chosen category instances in the image. \cite{Ranjan_2019_ICCV} shows the vulnerability of optical flow systems to adversarial patches. \cite{Subramanya_2019_ICCV} shows that adversarial patches can be created which fool network interpretation algorithms as well.

\noindent \textbf{Contextual reasoning in object detection:}
The relationship between objects and scene has been known for a long time to the computer vision community. Particularly, spatial context has been used in improving object detection. \cite{divvala2009empirical} empirically studies use of context in object detection.
\cite{torralba2003context} shows that scene classification and object classification can help each other. \cite{desai2011discriminative} utilizes spatial context in a structured SVM framework to improve object detection by detecting all objects in the scene jointly. \cite{heitz2008learning} learns to use stuff in the scene to find objects. \cite{oliva2007role,barnea2019exploring,rosenfeld2018elephant} discuss the role of context in object detection while \cite{shetty2019not} for classification and segmentation. \cite{Zhou2015ObjectDE} shows that a network trained for scene classification has implicitly learned object detection which suggests the inherent connection between scene classification and object detection.

More recently, with the emergence of deep networks, utilizing context has become easier and even unavoidable in some cases. For instance, Faster-RCNN, SSD, and YOLO process the input image only once to extract the features and then detect objects in the feature space. Since the features are coming from all over the image, the model naturally uses global features of the scene.

\section{Method}

\noindent {\bf Background on YOLO:} Given an image $x$, YOLO divides the image into a $S \times S$ grid where each grid cell predicts $B$ bounding boxes, so there are $BS^2$ possible bounding boxes. Assuming $C$ object classes, the model processes the image once and outputs, $P(object) \in \!R$: an objectness confidence score for each possible bounding box and $P(category|object) \in \!R^C$: scores of all categories conditioned on existence of an object at every grid cell, and a localization offset for each bounding box. During inference, the objectness score and class probabilities are multiplied to get the final score for each bounding box. In YOLO training, the objectness score is encouraged to be zero for background locations and closer to one for ground-truth object locations, and the class probabilities are encouraged to match the ground-truth only at the location of objects.

\noindent {\bf Adversarial patches:} Assume an image $x$, a recognition function $f(.)$, e.g., object classification, and a constant binary mask $m$ that is $1$ on the patch location and $0$ everywhere else. The mask covers a small region of the image, e.g., a box in the corner. We want to learn a noise patch $z$ that when pasted on the image fools the recognition function. Hence, in learning, we want to optimize:
$$z^* = \argmin_{z} \mathcal{L}\big(f(x \odot (1-m)+z \odot m; t\big)
\vspace{-0.05in}$$

\noindent where $\odot$ is the element-wise product, $t$ is the desired adversarial target, and $\mathcal{L}$ is a loss function that encourages the fooling of the model towards predicting the target. Note that any value in $z$ for which $m=\mathbf{0}$ will be ignored. 

In standard adversarial examples, we learn an additive perturbation so that when added to the input image, it fools a recognition function, e.g., object classifier or detector. Such a perturbation is bounded usually by $\ell_\infty$ norm to be invisible perceptually. However, in adversarial patches $z$ is not additive and is not bounded. It is only bounded to be in the range of allowed image pixel values in the mask location. This difference makes studying adversarial patches interesting since they are more practical (they can be used by printing and showing to the camera) and also are difficult to defend due to unconstrained perturbations.

\subsection{Our adversarial attacks:}

\noindent {\bf Per-image blindness attack:} We are interested in studying if an adversary can exploit contextual reasoning in object detection. Hence, given an image $x$ and an object category of interest $c$, we develop an adversarial patch that fools the object detector to be blind to category $c$ while the patch does not overlap with instances of $c$ on the image. 

Since we are interested in blindness, we should develop attacks that reduce the number of true positives rather than increasing false positives. We believe increasing the number of false positives is not an effective attack in real applications. For instance, in self-driving car applications, not detecting pedestrians can be more harmful than detecting many wrong pedestrians. Therefore, in designing our blindness attack, we do not attempt to fool the objectness score, $P(object)$ of YOLO and fool only the probability of the object category conditioned on being an object, $P(category|object)$.

We initialize our patch to be a black patch (all zeros). Then, we tune the noise $z$ to reduce the probability of the category $c$ that we want to attack on all locations of the grid that match the ground-truth. We do this by simply maximizing the summation of the cross-entropy loss of category $c$ at all those locations.
For optimization, we adopt a method similar to projected gradient descend (PGD) \cite{madry2018towards} in which after each optimization step and updating $z$, we project $z$ to be in the range of the acceptable pixel values [0-1] by clipping. Note that $z$ will have no contribution at the locations off the patch where $m=\mathbf{0}$. We stop the optimization when there is no detection for category $c$ on the image or we reach the maximum number of iterations.

\noindent {\bf Universal blindness attack:} Following \cite{moosavi2017universal}, we extend our attack to learn universal adversarial patches. For category $c$, we learn a universal patch $z$ on training data that makes the detector blind for category $c$ across unseen test images. To do so, we adopt the above optimization by iterating through training images in the optimization while keeping $z$ shared across all images.

\noindent {\bf Significance of blindness attack:} Note that there are two ways of reducing mAP in object detection: (1) introducing lots of false positives, (2) reducing true positives. We believe (1) can be achieved without exploiting the contextual reasoning by generating lots of false positives at the patch location to dominate the mAP calculation, which may not necessarily affect the true positives. However, since we are interested in showing the exploitation of contextual reasoning, we focus on (2) where the adversary should make the detector blind by reducing true positives.

To see this effect, we modify the mAP calculation in our experiments by removing any false positives at the patch location. We believe considering regular mAP for evaluation does not necessarily show the contextual exploitation.

\subsection{Defense for our adversarial attacks:}
\noindent Defending against adversarial examples has been shown to be challenging \cite{carlini2017adversarial,athalye2018obfuscated}. As discussed in the introduction, we believe defending against adversarial patches is even more challenging since the attack is expensive and the perturbation is not bounded to  lie in the neighborhood of the original image.

\noindent {\bf Grad-defense:}
\noindent Since we believe the main reason for the success of contextual adversarial patches is the exploitation of contextual reasoning by the adversary, we design our defense algorithm to limit the usage of contextual reasoning during training the object detector. 

In most fast object detectors including YOLO, each object location has a dedicated neuron in the final layer of the network, and since the network is deep, those neurons have very large receptive fields that span the whole image. To limit the usage of context by the detector, we are interested in limiting this receptive field only to the bounding box of the corresponding detection. 

One way of doing this is to hard-code a smaller receptive field by reducing the spatial size of the filters in the intermediate layers. However, this is not a great defense since: (1) it will reduce the capacity and thus accuracy of the model, (2) it shrinks the receptive field independent of the size of the detected box, thereby hurting the detection of large objects. We conducted an experiment where we change the network architecture of YOLOv2 and set the filter sizes for all layers after Layer16 (just before the pass-through connection) to 1x1. We observe that this model gives poor mAP on clean images which is reported in Table \ref{tab2}.

We believe that a better way of limiting the receptive field would be use a data-driven approach. Network interpretation tools like Grad-CAM \cite{selvaraju2016grad} that highlight the image regions which influence a particular network decision can be used. Grad-CAM works by visualizing the derivative of the output with respect to an intermediate convolutional layer (e.g., $conv5$ in AlexNet) that detects some high-level concepts. To limit the contextual reasoning in object detection, we should constrain such derivatives for a particular output to not span beyond the bounding box of the corresponding detected object. Hence, to defend against adversarial attacks, during YOLOv2 training, we calculate the derivative of each output with respect to an intermediate convolutional layer and penalize its nonzero values outside the detected bounding box.

More formally, assuming $y^c$ is the confidence of an object belonging to category $c$ detected at bounding box $B$ and $A_{ij}^k$ for the activation of a convolutional layer at location $(i,j)$ and channel $k$, we calculate the derivative $\frac{\partial y^c}{\partial A_{ij}^k}$ and normalize it to sum to $1$ over the whole feature map: 
\vspace{-.1in}
\begin{equation} \label{eq:beta_eq}
\hat{\beta}_{ij} = \frac{\beta_{ij}}{\sum_{i,j} {\beta_{ij}}} \quad \text{where} \quad \beta_{ij} = \sum_k { \bigg|\frac{\partial y^c}{\partial A_{ij}^k}\bigg|}
\end{equation}
\vspace{-.1in}

Then, we minimize the following loss to encourage  $\hat\beta$ values to be completely inside the bounding box $B$.
\vspace{-.1in}
\begin{equation} \label{eq:Loss_eq}
    \mathcal{L} = -\sum_{{i,j}\in B} \hat\beta_{ij}
\end{equation}
\vspace{-.1in}

Since $\hat\beta$ sums to $1$, minimizing this loss will minimize the influence of image regions outside the detected bounding box on its corresponding object detection. We believe this is a regularizer that limits the receptive field of the final layer depending on the size of detected objects. So, it should limit the contextual reasoning of the object detector. Interestingly, this loss can be even minimized on unlabeled data in a semi-supervised setting.

\begin{figure}[t]
    \centering
    \deflen{mylengtha}{0.4\linewidth}
    \begin{subfigure}[b]{\mylengtha}
        \includegraphics[width=\linewidth]{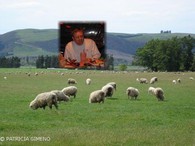}
        \caption*{}
    \end{subfigure}
    \begin{subfigure}[b]{\mylengtha}
        \includegraphics[width=\linewidth]{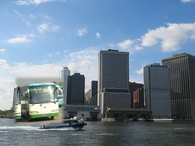}
        \caption*{}
    \end{subfigure}
    \vspace{-0.2in}
    \caption{\textbf{Out of context images-} This figure shows examples
    from the out of context dataset we curated to train our OOC defense model.
     }
     \label{fig:ooc}
     \vspace{-0.2in}
\end{figure}

\noindent {\bf Out-of-context(OOC) defense:} Another way of limiting contextual reasoning is to remove the influence of context from the training data. We do so by simply overlaying an out-of-context foreground object on the training images. To create the dataset, we take two random images from the PASCAL VOC training data, crop one of the annotated objects from the first image and paste it at the same location on the second image. We blur the boundary of the pasted object to remove sharp edges and also remove the annotations of the second image corresponding to the objects occluded by the added foreground object. We keep non-overlapping annotations intact. We train YOLOv2 %
on the new dataset to get a model that is less dependent on context. Fig. \ref{fig:ooc} shows example out of context training images. A few other defense algorithms used as baselines are described in the experiments section.

\section{Experiments}

\noindent We use PASCAL VOC dataset \cite{Everingham15} for all our experiments \footnote{Code available at \url{https://github.com/UMBCvision/Contextual-Adversarial-Patches}}. For each chosen attack category, we create a subset of images which contain the object category and where none of the  instances overlap with the patch area. This results in a filtered image set per category. We also evaluate the universal blindness attack on KITTI dataset \cite{Geiger2012CVPR} for ``pedestrian'' and ``car''.

\subsection {Implementation details:}
\noindent We use Pytorch \cite{paszke2019pytorch} implementation of YOLOv2 and NVIDIA Titan X GPUs to run 
our experiments. The image size is fixed at $416\times416$ and our patch size is fixed at $100$x$100$ which covers less than $6\%$ of the image area. The top-left corner of the patch is at pixel $(5, 5)$. When optimizing, we initialize $z$ from all zeros (black patch) and use the Adam optimizer. We also experimented with the standard $l_\infty$ PGD attack but it did not provide any improvement in attack performance or convergence. For per-image experiments, we use the learning rate of $0.1$ and run the attack for 250 iterations per image. For universal patch experiments, we use learning rate of $0.01$, minibatch size of $1$, $10$ iterations per minibatch, and total of $100$ epochs. The confidence threshold, NMS threshold, and IOU overlap threshold used for evaluations are $0.005$, $0.45$ and $0.5$ respectively. 

For Grad-defense experiments, we obtained best results considering the gradients at 16th layer of YOLOv2 (just before the pass-through connection), and $\lambda=1.0$ which balances the YOLO standard loss and our loss in Eq. \ref{eq:Loss_eq}. 

\begin{table*}[!ht]
    \centering
    \scalebox{0.61}{
    \begin{tabular}{||c||c|cccccccccccccccccccc||}
        \hline
              & \small{Mean} & \small{aero}  & \small{bike}  & \small{bird} & \small{boat} & \small{bottle} & \small{bus} & \small{car} & \small{cat} & \small{chair}     & \small{cow} & \small{dtable} & \small{dog} & \small{horse}
              & \small{mbike} & \small{person} & \small{pplant}
              & \small{sheep} & \small{sofa} & \small{train}
              & \small{tv} \\ \hline
        \small{Total No. images } &- &205 &250 &289 &176 &240 &183 &775 &332 &545 &127 &247 &433 &279 &233 &2097 &254 &98 &355 &259 &255  \\
        \small{No. filtered images} &- &136 &190 &182 &114 &205 &102 &510 &160 &454 &87 &212 &244 &173 &142 &1286 &174 &73 &273 &148 &187     \\ 
    \hline
        \small{YOLOv2 (clean)} &{\bf 76.04} &75.05 &81.02 &75.22 &66.58 &50.59 &81.08 &79.86 &80.96 &64.40 &85.19 &76.32 &85.35 &85.91 &80.08 &75.62 &57.28 &79.90 &79.83 &83.30 &77.18   \\
        \small{White patch} 
        &76.33&75.30&80.59&76.01&67.00&50.69&81.20&79.85&80.89&64.18&84.83&76.99&85.93&86.33&80.23&75.69&57.75&79.86&81.02&84.82&77.42\\
        \small{Random noise patch} &76.20&75.00&80.63&75.87&66.29&50.24&81.07&79.60&81.07&64.23&85.37&76.96&86.34&86.28&79.81&75.60&57.17&80.03&80.71&84.78&76.86\\
        \small{OOC attack}
        &75.93&74.46&80.40&75.56&65.02&50.83&81.61&79.58&81.01&63.61&84.74&76.92&86.25&86.42&79.39&75.68&56.32&79.82&80.83&82.89&77.29\\
        \small{\bf Adv patch attack (Ours) 
        } &{\bf 55.42}&40.89&71.51&44.11&38.46&39.90&60.25&62.28&57.25&54.33&54.03&71.27&62.90&67.98&66.77&59.87&38.48&55.53&64.14&47.96&50.56\\
    \hline        
        \small{Grad-Defense (clean)} &\textbf{75.83}&73.41&80.71&75.33&61.32&47.28&81.33&81.05&84.19&63.11&86.02&77.01&86.14&85.91&82.84&74.39&53.34&81.01&81.12&84.73&76.40\\
        \small{Grad-Defense (attacked)} &{\bf 61.31}&51.71&75.53&53.17&42.88&40.20&66.96&71.43&62.13&56.29&59.59&73.46&69.91&75.58&74.98&63.32&40.23&59.47&70.86&61.95&56.45\\
        \hline
    \end{tabular}
    }
    \vspace{-0.1in}
    \caption{\textbf{Per-image blindness attacks, and defense evaluation-} ``YOLOv2 (clean)'' shows the results of YOLOv2 on our clean per-image attack test sets. AP is calculated only on the filtered subset of images. ``Adv patch attack'' is our adversarial patch algorithm. We also evaluate three baseline attacks. Qualitative examples are shown in Fig.  \ref{fig:perimagepatchfigure}.}
    \label{tab1}
\end{table*}

\begin{table*}[!ht]
    \centering
    \scalebox{0.59}{
    \begin{tabular}{||c||c|cccccccccccccccccccc||}\hline
          & \small{Mean} & \small{aero}  & \small{bike}  & \small{bird} & \small{boat} & \small{bottle} & \small{bus} & \small{car} & \small{cat} & \small{chair}     & \small{cow} & \small{dtable} & \small{dog} & \small{horse}
          & \small{mbike} & \small{person} & \small{pplant}
          & \small{sheep} & \small{sofa} & \small{train}
          & \small{tv} \\ \hline
        \small{Total No. training imgs} &- &68 &95&91&57&102&51&255&80&227&43&106&122&86&71&643&87&36&136&74&93    \\ 
        \small{Total No. testing imgs} &- &68 &95&91&57&102&51&255&80&227&43&106&122&86&71&643&87&36&136&74&93    \\ 

    \hline
        \small{YOLOv2 (clean)} &{\bf 76.85} &79.25&83.17&77.19&63.88&49.70&80.61&79.47&80.59&64.92&85.76&77.39&86.65&81.32&84.78&75.41&56.82&89.05&76.96&87.59&76.56 \\
        \small{\bf YOLOv2 (attacked) (Ours)}&{\bf 56.24}&29.66&71.51&39.7&34.14&44.67&65.21&60.26&44.41&58.28&61.94&77.12&67.52&67.82&59.16&65.2&46.17&69.87&72.04&42.07&47.96\\
    \hline
        \small{AT-2000 (clean)}&64.01&53.17&69.84&58.82&41.98&39.34&67.42&71.15&70.74&50.27&69.97&71.35&77.09&77.06&75.72&69.16&44.89&76.06&63.26&62.05&70.95\\
        \small{AT-2000 (attacked)}&41.55&17.36&55.57&27.36&17.55&36.41&31.67&47.02&39.34&42.76&38.98&71.93&64.01&38.12&51.38&57.16&28.82&42.55&59.88&22.70&40.43\\
    \hline        
        \small{AT-30 (clean)}&70.47&73.81&75.60&65.49&54.73&43.28&76.81&75.87&71.15&57.14&81.41&74.04&72.66&77.81&79.42&68.89&47.10&85.65&74.79&78.93&74.80\\
        \small{AT-30 (attacked)}&50.47&29.23&75.83&27.16&42.23&38.86&46.15&65.27&41.22&51.20&61.10&75.73&46.88&40.40&76.21&53.05&37.21&60.19&56.96&31.00&53.42\\		
    \hline        
        \small{OOC Defense (clean)} &65.67&79.93&75.63&59.53&48.54&34.68&78.39&74.23&71.35&44.51&68.22&76.31&69.73&72.50&68.18&66.02&38.90&74.34&64.55&76.84&70.93 \\
        \small{OOC Defense (attacked)} &60.35&70.81&70.03&45.84&44.47&33.06&76.87&70.98&60.26&36.89&67.55&76.01&64.47&71.40&61.63&64.26&37.42&68.68&63.72&63.98&58.75\\
    \hline                
        \small{YOLOv2 1x1 (clean)} &59.55&69.34&68.79&51.15&41.41&43.48&61.86&73.98&51.71&48.28&72.96&55.08&66.58&65.96&70.29&66.01&41.00&77.89&43.04&54.52&67.59 \\
        \small{YOLOv2 1x1  (attacked)} &59.57&69.25&68.84&51.07&41.13&43.67&61.66&74.12&52.09&48.46&72.89&55.11&66.34&65.96&71.02&65.85&41.36&76.67&42.95&54.54&68.42\\        
    \hline                
        \small{Gradient w.r.t. input (clean)} &65.80&62.18&75.85&58.91&46.55&31.78&76.12&62.96&72.72&53.83&68.74&73.16&69.46&76.41&75.47&58.32&62.30&74.45&73.06&72.20&71.51\\
        \small{Gradient w.r.t. input (attacked)} &48.97&22.39&67.68&29.68&22.02&29.30&67.70&45.03&39.96&49.19&42.80&74.71&61.23&55.89&65.10&46.45&41.66&51.78&69.53&42.39&54.82	\\        
    \hline        
        \small{Grad-Defense (clean)} &\textbf{76.09}&76.31&83.08&75.49&65.15&46.41&83.11&80.53&81.99&63.73&87.22&77.63&86.4&81.26&84.57&74.12&51.69&83.88&78.26&85.95&74.94\\
        \small{Grad-Defense (attacked)} &{\bf 64.84}&50.9&80.47&50.11&48.00&46.23&78.93&68.83&63.11&60.13&67.96&77.62&77.26&77.67&68.92&67.05&48.50&70.99&73.97&60.3&59.79\\
        \hline
    \end{tabular}
    }
    \vspace{-0.1in}
        \caption{\textbf{Universal blindness attacks, and defense evaluation-} ``YOLOv2 (clean)'' shows the results of YOLOv2 on our clean universal attack test sets. Note that the image sets are different from Table \ref{tab1}. ``YOLOv2 (attacked)'' shows our universal patch blindness attack. We report results for two adversarially trained models (2000 and 30 iterations). ``OOC Defense'' refers to YOLOv2 trained on our out-of-context dataset. For YOLOv2 1x1, we limit the size of final conv layers to 1x1.  ``Gradient w.r.t. input'' penalizes the gradient of the output w.r.t the input in pixel space. ``Grad-Defense'' is our main defense algorithm that outperforms other algorithms on both clean and attacked images. Qualitative examples are shown in Fig. \ref{fig:universalpatchfigure}.
    }
\label{tab2}
\end{table*}

\subsection{Evaluation:} 
\noindent Since we are interested in learning adversarial patches that do not overlap with the object, we fix the location of the patch, e.g., at top-left corner, and run evaluations only for images on which there is no object of interest overlapping with the patch location as described earlier.

Because our filtered image sets are independent, we run evaluations independently for each category and report the AP scores for the category in question. Hence, the mean average precision (mAP) we are reporting is a little different from common practice since it is mean over categories that use different image sets. This is due to the fact that there can be multiple objects in an image and such images can be part of different image sets. As mentioned earlier, we remove the false positives overlapping with the patch because we do not want our attack to work by making the patch as the most salient object. 

\begin{figure}[!h]
\captionsetup{font=small}
  \begin{center}
  \begin{tabular}{|c c|}
\hline  
 \footnotesize{YOLOv2}  & \footnotesize YOLOv2 \\
\footnotesize Detection &  \footnotesize Adv patch Detection \\
\hline
\vspace{-.08in}
&\\
\vspace{-.05in}
\begin{sideways} \scriptsize ~~~~~~~~~~~~~Target: Person \end{sideways}
\includegraphics[width=.15\textwidth]{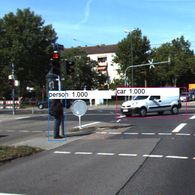}&

\includegraphics[width=.15\textwidth]{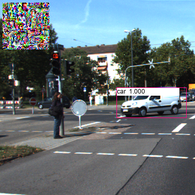}\\
& \small Person fooled \\
&\\

\vspace{-.05in}
\begin{sideways} \scriptsize ~~~~~~~~~~~~~Target: Car \end{sideways}
\includegraphics[width=.15\textwidth]{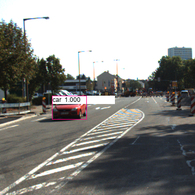}&

\includegraphics[width=.15\textwidth]{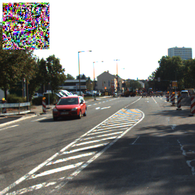}\\
& \small Car fooled \\

\hline
\end{tabular}
  \caption{\textbf{Evaluation on KITTI dataset-} Examples of trained universal patch for car and person.}
\label{fig:kitti}
  \end{center}
  \vspace{-0.4in}
\end{figure}
\subsection{Results and discussion}
\noindent In this section, we evaluate our attacks and defense algorithms in various settings. 

\noindent {\bf Per-image blindness attack:} We consider three baselines where we replace the adversarial patch with all white patch, an i.i.d. noise patch, or an out-of-context patch and show that our learned adversarial patch is better.

Table \ref{tab1} shows the results of our per-image blindness attack experiments. We show the total number of images in PASCAL VOC 2007 and our filtered images. The ``White patch'' and ``Random noise patch'' baselines do not reduce the accuracy which means the attack needs to be tuned. Our attack reduces the mAP from $76.04\%$ to $55.42\%$ which is $20.61$ points reduction. The highest drop of almost $35$ points is for ``train'' and ``aeroplane''. 

\noindent {\bf Universal blindness attack:} Table \ref{tab2} shows the results of our universal blindness attack experiments. We divide the filtered images to two halves for training and testing and then learn a universal adversarial patch per category on the training set and test it on the testing set.  We see that our attack reduces the mAP for YOLOv2 from $76.85\%$ to $56.24\%$ which is almost $20.5$ points reduction in mAP. 

\noindent {\bf Defense evaluations:} The results of defense algorithms are reported in Table \ref{tab2}. Our Grad-defense algorithm outperforms other methods in terms of both accuracy on clean images ($76.09\%$) and attacked images ($64.84\%$). The original YOLOv2 model has an accuracy of $76.85\%$ on clean images and $56.24\%$ on attacked images. This shows the effectiveness of our defense method on limiting the contextual reasoning. In Table \ref{tab2}, we also evaluate a few other defense algorithms including training with out of context data. Our Grad-defense is also effective for per-image attacks. The results are reported in Table \ref{tab1}.

\noindent {\bf Adversarial training:} Adversarial training is a common defense method for regular adversarial examples. Unfortunately, this is not suitable for adversarial patches since calculating each adversarial patch example is computationally expensive. For completeness, we use this as a baseline. We use 30 and 2000 iterations to generate adversarial patch examples and use them in training the model. The results are reported in Table \ref{tab2}. Even though we run the 2000 iteration attack for almost 6 days, its accuracy on clean images was still very low. The adversarially trained model has a clean accuracy of $70.47\%$ but it has poor performance after attack and gives $50.47\%$ accuracy.

\noindent {\bf Patch detection and Gaussian blurring: }One might wonder that because the adversarial patches are not norm-constrained and look noisy, it should be easy to detect their presence on the image using a simple network or blur them using a Gaussian filter. In our experiments, we show that by incorporating these techniques into our patch training loop, we can still fool the patch detection or blurring system. 
\begin{table*}[!ht]
    \centering
    \scalebox{0.57}{
    \begin{tabular}{||c||c|cccccccccccccccccccc||}\hline
          & \small{Mean} & \small{aero}  & \small{bike}  & \small{bird} & \small{boat} & \small{bottle} & \small{bus} & \small{car} & \small{cat} & \small{chair}     & \small{cow} & \small{dtable} & \small{dog} & \small{horse}
          & \small{mbike} & \small{person} & \small{pplant}
          & \small{sheep} & \small{sofa} & \small{train}
          & \small{tv}  \\ \hline
        \small{YOLOv2 (attacked)}
        &56.24&29.66&71.51&39.7&34.14&44.67&65.21&60.26&44.41&58.28&61.94&77.12&67.52&67.82&59.16&65.2&46.17&69.87&72.04&42.07&47.96\\
        \small{Patch detector accuracy} 
        &{\bf 98.27}&100.00&95.79&100.00&92.98&100.00&100.00&98.43&98.75&99.12&100.00&91.51&96.72&100.00&95.77&98.44&100.00&100.00&97.81&100.00&100.00\\
    \hline
        \small{YOLOv2 (fool detector)}
        &57.66&33.17&67.18&39.30&34.27&44.47&77.57&60.49&51.15&57.58&61.91&77.40&76.05&68.39&60.18&66.97&46.41&63.29&67.97&42.56&56.91 \\
        \small{Patch detector accuracy} &{\bf 5.54}&2.94&4.21&7.69&1.75&0.97&25.49&1.57&18.75&1.76&11.36&0.00&4.10&2.30&2.82&4.98&6.90&2.70&0.00&4.05&6.38 \\
        \hline
    \end{tabular}
    }
    \vspace{-0.1in}
    \caption{\textbf{Fooling patch detector-} We show that even though a norm-unconstrained adversarial patch can be detected because of its high frequency components, such a detector can be incorporated into the training loop and be fooled by our universal blindness attack with similar efficiency.}
    \label{tab:Fool patch detector}
    \vspace{-0.15in}
\label{tab4}
\end{table*}

For the patch detection network, we create a dataset of clean images and attacked images and finetune a ResNet-18 network for a binary classification task - whether a patch is present in an image or not. We evaluate the performance of the patch detector on our attacked images. Then, we incorporate the patch detector into our adversarial patch optimization loop by adding another loss which when minimized fools the patch detector. We see that using the same attack setting, we are able to achieve comparable attack efficiencies and are now able to fool the patch detector as well. The results are shown in Table \ref{tab4}. We believe one can still train another detector, so a more advanced defense would be to train the attack and detector together in an adversarial game setting. However, we believe it is beyond the scope of this paper.
For the Gaussian blurring case, we create universal adversarial patches using our method and add that to the image. The resultant image is blurred using a 7x7 Gaussian kernel to limit the effectiveness of the patch. We observed that after blurring, we see an increase of mAP from $56.24\%$ to $71.53\%$. This means that the blurring reduced the effectiveness of our patch.
A similar idea was discussed as a form of defense for $\ell_\infty$ adversarial examples in \cite{li2017adversarial}. However, it was shown in \cite{carlini2017adversarial} that such a defense can be overcome by creating the adversary with the knowledge of blurring. We perform a similar experiment to show that adversarial patches can be created using the knowledge of blurring filter. We include the Gaussian filter as an additional layer at the input and train our universal patch iteratively by applying the blurring filter at randomly selected iterations. After attack, this gives us an mAP of $53.25\%$ which is comparable to the results that we had obtained without incorporating blurring during training. 

\ignore{
\begin{table}[h]
    \centering
    \scalebox{1}{
    \begin{tabular}{@{}lcc@{}}\toprule
                              & \small{car}  & \small{person}  \\ \midrule
        \small{YOLO (clean)} & 59.0 & 51.7 \\
        \small{YOLO (attacked)} & 31.8 & 34.2\\
     \bottomrule   
\end{tabular}
    }
    \caption{\textbf{KITTI evaluation} We evaluate the performance
    of a YOLO model pretrained on VOC on our dataset curated from KITTI and report the AP scores per class. }
\label{tab3}
\end{table}
}

\noindent {\bf Per-image targeted attack:} Though our main objective is to show the presence of blindness attack, we also evaluate a targeted attack. For each target object category, we choose 500 random images that do not contain the object. Then, we ``artificially'' change all ground-truth object labels to the target category in evaluation. We expect our targeted attack to increase the mAP on this artificial ground-truth from zero to a higher number. We use learning rate of 0.1 and run the attack for 2,000 iterations per image. On evaluation after attack, we see that the mAP increases from 0 to $18.61\%$. The detailed results are reported in the appendix.

\noindent {\bf Per-image objectness attack:} Because YOLO makes its decision based on objectness as well as class scores, it is interesting to see whether we can fool objectness. This can be seen as a different form of attack. It is important to note that all mAP evaluations are done keeping the objectness threshold for acceptance at the original value of 0.005. To perform objectness attack, we run the patch optimization for 2,000 iterations per image and encourage the objectness scores to be below 0.005. We use learning rate of 0.05 for this experiment. We see a $17.5$ point drop in mAP from $76.04\%$ to $58.49\%$. The detailed results are provided in the appendix.

\noindent {\bf Universal blindness attack for KITTI:}  We evaluate our universal blindness attack on person (pedestrian) and car detection on KITTI dataset which is closer to self-driving car applications. Similar to PASCAL VOC experiments, we filter out the images for which the objects of interest overlap with the patch, but since the patch is on the top-left corner and people and cars are usually on the ground, we see that it mostly never overlaps with the objects of interest. Then we use the YOLOv2 model trained on PASCAL VOC dataset and learn universal adversarial patches for person and car categories. Since YOLOv2 needs to resize the image to be square and KITTI frames have a different aspect ratio, we crop a random square box from the middle of KITTI frames so that we do not degrade the image quality by stretching it. Qualitative results are shown in Fig. \ref{fig:kitti}. Our attack reduces car detection AP from $59.0\%$ to $31.8\%$ and person detection from $51.7\%$ to $34.2\%$ which are significant reductions. 

\begin{figure*}[!h]
\captionsetup{font=small}
  \begin{center}
  \begin{tabular}{| c c c c |}
\hline  
 \footnotesize{YOLOv2} &  \footnotesize \footnotesize{YOLOv2} & \footnotesize Grad-Defense & \footnotesize Grad-Defense \\
\footnotesize Detection & \footnotesize Adv patch Detection & \footnotesize{Detection} &  \footnotesize Adv patch Detection \\
\hline
\vspace{-.08in}
&&&\\
\vspace{-.05in}
\begin{sideways} \scriptsize ~~~~~~~~~~~~~Target: Person \end{sideways}
\includegraphics[width=.14\textwidth]{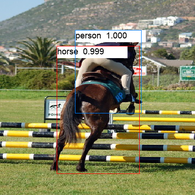}&

\includegraphics[width=.14\textwidth]{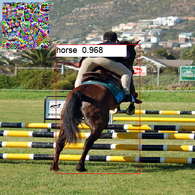}&
\includegraphics[width=.14\textwidth]{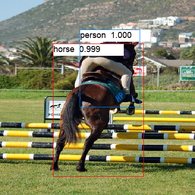}&
\includegraphics[width=.14\textwidth]{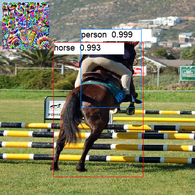}\\
& \small Person fooled && \small Person detected\\
&&&\\

\vspace{-.05in}
\begin{sideways} \scriptsize ~~~~~~~~~~~~~Target: Dog \end{sideways}
\includegraphics[width=.14\textwidth]{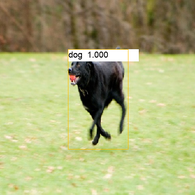}&

\includegraphics[width=.14\textwidth]{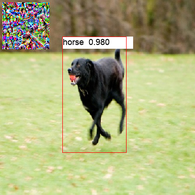}&
\includegraphics[width=.14\textwidth]{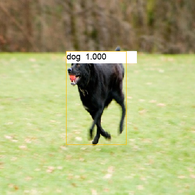}&
\includegraphics[width=.14\textwidth]{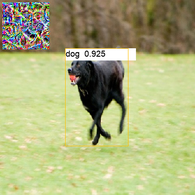}\\
& \small Dog fooled & & \small Dog detected\\
\hline
\end{tabular}
\vspace{-.1in}
  \caption{\textbf{Universal patch blindness attack and Grad-Defense results-} We compare the detection results of YOLOv2 with our Grad-Defense model . We attack each model separately. We see that our Grad-Defense model is less susceptible to context-based attack. The target category is written below each example. The universal patch per category is always on the top-left corner.}
\label{fig:universalpatchfigure}
  \end{center}
   \vspace{-.1in}
\end{figure*}
\begin{figure*}[!h]
 
\vspace{-.1in}
\captionsetup{font=small}
  \begin{center}
  \begin{tabular}{| c c c c c c|}
\hline  \footnotesize YOLOv2 & \footnotesize{YOLOv2} & \footnotesize{YOLOv2} & \footnotesize Grad-defense & \footnotesize Grad-defense & \footnotesize Grad-defense \\
\footnotesize{Detection} & \footnotesize Grad-CAM \cite{selvaraju2016grad}& \footnotesize DetGrad-CAM & \footnotesize{Detection} & \footnotesize Grad-CAM \cite{selvaraju2016grad} & \footnotesize DetGrad-CAM \\
\hline
\vspace{-.08in}
&&&&&\\
\vspace{-.05in}
\begin{sideways} \scriptsize ~~~~Category: Boat \end{sideways}
\includegraphics[width=.13\textwidth]{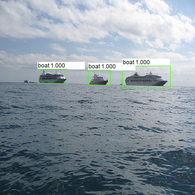}&
\includegraphics[width=.13\textwidth]{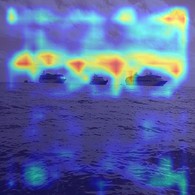}&
\includegraphics[width=.13\textwidth]{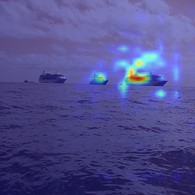}&
\includegraphics[width=.13\textwidth]{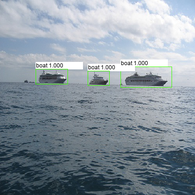}&
\includegraphics[width=.13\textwidth]{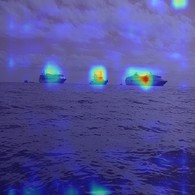}&
\includegraphics[width=.13\textwidth]{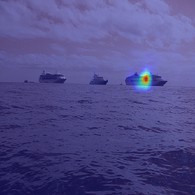}\\
&&&&&\\

\vspace{-.05in}
\begin{sideways}  \scriptsize ~~~~Category: Horse \end{sideways}
\includegraphics[width=.13\textwidth]{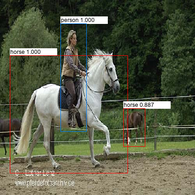}&
\includegraphics[width=.13\textwidth]{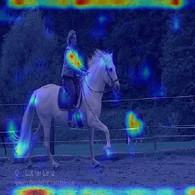}&
\includegraphics[width=.13\textwidth]{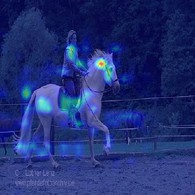}&
\includegraphics[width=.13\textwidth]{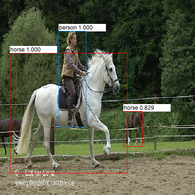}&
\includegraphics[width=.13\textwidth]{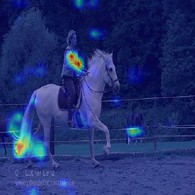}&
\includegraphics[width=.13\textwidth]{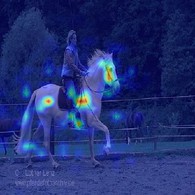}\\
&&&&&\\
\hline
\end{tabular}
\vspace{-.1in}
  \caption{\textbf{DetGrad-CAM-} We observe that modifying Grad-CAM gives us better visual interpretations for the object detector. For each image we show the interpretation results of a detected object using Grad-CAM\cite{selvaraju2016grad} and DetGrad-CAM from YOLOv2 and also our Grad-defense model. The category for which interpretation is calculated is written on the left. Note that for the first row, we are interested in finding the interpretation for the right-most boat and interestingly Grad-CAM highlights all three boats as important for detection.}
 
\vspace{-.35in}
  
\label{fig_detgradcam}
  \end{center}
\end{figure*}
\subsection{DetGrad-CAM for visualization of context:}
To analyze the usage of contextual information in object detection, we may use interpretation algorithms similar to those in object classification literature \cite{selvaraju2016grad,simonyan2013deep,zhou2015cnnlocalization}. We consider a popular algorithm, Grad-CAM \cite{selvaraju2016grad} which is for classification tasks. Specifically, we want to highlight the regions in the image responsible for the detection of each object. We can then analyze the amount of contextual information used by the model.
To summarize the Grad-CAM algorithm, we use notations defined for Equation \ref{eq:beta_eq}. The algorithm measures the importance score of each intermediate feature on the confidence $y^c$ of an object by calculating: 
\vspace{-0.075in}
\begin{equation} \label{eq:alpha_eq}
\alpha^c_k = \frac{1}{Z}\sum_{i,j}{\frac{\partial y^c}{\partial A^k_{ij}}}
\vspace{-0.075in}
\end{equation}
where $Z$ is a normalizer. Then we calculate the weighted summation of activations and consider only the positive values, i.e, $G_{ij}^c = max(0, \sum_k\alpha^c_kA^k_{ij})$. The final heatmap  $G_{ij}^c$ is resized to be in the size of input image. 

However, the summation in Equation \ref{eq:alpha_eq} has the effect of removing the spatial information when calculating the importance score. If an image has multiple objects of the same category, this can lead to highlighting all object instances for each detected instance. Since object detectors are trained to classify as well as localize the objects, the gradient values of the interpretation algorithm contain information regarding the location of the object. As a result, summation over all locations would result in loss of the spatial information. Hence, we modify the algorithm for detection task, namely \textbf{DetGrad-CAM}, by doing element-wise multiplication of gradients with the activations, and then summing over all features, i.e we calculate the heatmap as follows:
\vspace{-0.10in}
\begin{equation} \label{eq:detgradcam}
\tilde G_{ij}^c = max(0, \sum_k\frac{\partial y^c}{\partial A^k_{ij}} \odot A^k_{ij})
\end{equation}

We believe this simple modified method results in better explanation for object detection models as shown in Fig. \ref{fig_detgradcam}. We also see that the Grad-defense model is using lesser contextual information compared to original model which leads to better defense against context-based adversaries.  

\newpage
\section{Conclusion}
\vspace{-0.1in}
\noindent Object detectors like YOLO, SSD and Faster-RCNN have become popular due to their fast inference time. They process the image only once rather than running the model on every proposal e.g., in RCNN. Hence, such models naturally learn to employ contextual reasoning which results in better accuracy. In this paper we show that reliance on context makes the detector vulnerable to category specific adversarial patches in which the adversary can make the detector blind for a category chosen by the attacker without occluding those objects in the scene or affecting the detection of other categories. This is also a practical attack that can cause major issues when deep models are deployed in real world applications like self-driving cars. We propose a defense algorithm by regularizing the model to limit the influence of image regions outside the bounding boxes of the detected objects. We show that the model trained this way improves robustness to the contextual attack. We believe this highlights the need for studying defense algorithms which are robust to contextual adversarial patch attacks but also accurate.

\noindent {\bf Acknowledgement:} This work was performed under the following financial assistance award: 60NANB18D279 from U.S. Department of Commerce, National Institute of Standards and Technology, funding from SAP SE, and also NSF grant 1845216.

\clearpage
{\small
\bibliographystyle{ieee}
\bibliography{ms.bib}
}

\onecolumn
\pagebreak
\begin{appendices}

\renewcommand{\thefigure}{A\arabic{figure}}
\renewcommand{\thetable}{A\arabic{table}}

\setcounter{figure}{0}
\setcounter{table}{0}

We show more results in the appendix. Please refer to the captions of tables and figures for the description.
 
\begin{figure}[h!]
\centering
\includegraphics[width=.25\textwidth]{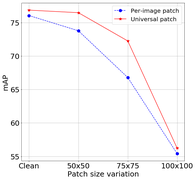}
\caption{{\bf Sensitivity to patch size -} We study the effect of the variation of patch size on our blindness attacks. We observe that as the patch size decreases, the attack success rate decreases. These results are also shown in Table \ref{taba1:Per image patch patch size ablation}.}
\label{figa1:ablationpatchsize}
\end{figure}

\begin{table*}[h!]
    \centering
    \scalebox{0.6}{
    \begin{tabular}{||c||c|cccccccccccccccccccc||}\hline
          & \small{Mean} & \small{aero}  & \small{bike}  & \small{bird} & \small{boat} & \small{bottle} & \small{bus} & \small{car} & \small{cat} & \small{chair}     & \small{cow} & \small{dtable} & \small{dog} & \small{horse}
          & \small{mbike} & \small{person} & \small{pplant}
          & \small{sheep} & \small{sofa} & \small{train}
          & \small{tv}  \\ \hline
        {\bf Per-image} &&&&&&&&&&&&&&&&&&&&&\\
        \small{YOLOv2 (clean)}
        &76.04 &75.05 &81.02 &75.22 &66.58 &50.59 &81.08 &79.86 &80.96 &64.40 &85.19 &76.32 &85.35 &85.91 &80.08 &75.62 &57.28 &79.90 &79.83 &83.30 &77.18\\
        \small{YOLOv2 (50x50 patch)} 
        &73.77&72.20&80.36&72.95&61.18&48.99&80.55&78.55&80.73&62.94&77.78&75.67&84.34&81.09&78.81&74.54&53.19&77.12&79.03&82.03&73.27\\
        \small{YOLOv2 (75x75 patch)} 
        &66.79&61.30&76.79&62.69&48.42&46.31&72.39&72.68&69.90&59.59&65.88&75.06&77.06&80.35&76.81&70.35&44.55&68.11&75.88&69.68&61.97\\
        \small{YOLOv2 (100x100 patch)}
        &55.42&40.89&71.51&44.11&38.46&39.90&60.25&62.28&57.25&54.33&54.03&71.27&62.90&67.98&66.77&59.87&38.48&55.53&64.14&47.96&50.56\\
        \hline
        {\bf Universal} &&&&&&&&&&&&&&&&&&&&&\\
        \small{YOLOv2 (clean)} 
        &76.85&79.25&83.17&77.19&63.88&49.70&80.61&79.47&80.59&64.92&85.76&77.39&86.65&81.32&84.78&75.41&56.82&89.05&76.96&87.59&76.56 \\
        \small{YOLOv2 (50x50 patch)} 
        &76.47&78.86&82.35&77.39&62.01&49.96&80.71&78.48&80.51&64.72&84.12&78.18&86.83&80.77&84.68&75.73&56.93&86.43&75.86&87.49&77.47\\
        \small{YOLOv2 (75x75 patch)} 
        &72.25&66.48&82.88&73.85&59.93&45.79&80.42&77.08&68.55&62.07&74.12&77.62&78.63&80.94&77.02&75.09&53.09&80.64&75.09&85.04&70.57\\
        \small{YOLOv2 (100x100 patch)}
        &56.24&29.66&71.51&39.7&34.14&44.67&65.21&60.26&44.41&58.28&61.94&77.12&67.52&67.82&59.16&65.2&46.17&69.87&72.04&42.07&47.96 \\
        \hline
    \end{tabular}
    }
    \caption{{\bf Sensitivity to patch size -} The first 4 rows are the per-image blindness attack and the last 4 rows are the universal blindness attack.}
    \label{taba1:Per image patch patch size ablation}
\label{tab6}
\end{table*}

\begin{table*}[!ht]
    \centering
    \scalebox{0.61}{
    \begin{tabular}{||c||c|cccccccccccccccccccc||}\hline
          & \small{Mean} & \small{aero}  & \small{bike}  & \small{bird} & \small{boat} & \small{bottle} & \small{bus} & \small{car} & \small{cat} & \small{chair}     & \small{cow} & \small{dtable} & \small{dog} & \small{horse}
          & \small{mbike} & \small{person} & \small{pplant}
          & \small{sheep} & \small{sofa} & \small{train}
          & \small{tv}  \\ \hline
        \small{YOLOv2 (clean)}&76.04 &75.05 &81.02 &75.22 &66.58 &50.59 &81.08 &79.86 &80.96 &64.40 &85.19 &76.32 &85.35 &85.91 &80.08 &75.62 &57.28 &79.90 &79.83 &83.30 &77.18    \\
        \small{YOLOv2 (attacked)} &58.49&38.60&73.86&49.39&34.27&41.12&60.50&61.50&71.91&53.38&56.74&73.24&71.94&75.01&70.69&58.30&38.48&58.99&70.61&58.23&52.98\\
        \hline
    \end{tabular}
    }
    \caption{\textbf{Per-image objectness attack -} This attack is described in \textbf{Per-image objectness attack} paragraph of Section 4.3 in the main paper and the qualitative results are in Figure \ref{figa4:objectnessfigure} of the appendix. We perform a different kind of adversarial patch attack by trying to fool YOLOv2 objectness confidence.}
    \label{tab:Objectness attack}
\label{tab5}
\end{table*}

\begin{table*}[!ht]
    \centering
    \scalebox{0.63}{
    \begin{tabular}{||c||c|cccccccccccccccccccc||}\hline
          & \small{Mean} & \small{aero}  & \small{bike}  & \small{bird} & \small{boat} & \small{bottle} & \small{bus} & \small{car} & \small{cat} & \small{chair}     & \small{cow} & \small{dtable} & \small{dog} & \small{horse}
          & \small{mbike} & \small{person} & \small{pplant}
          & \small{sheep} & \small{sofa} & \small{train}
          & \small{tv}  \\ \hline
        \small{Clean} 
        &0.0&0.0&0.0&0.0&0.0&0.0&0.0&0.0&0.0&0.0&0.0&0.0&0.0&0.0&0.0&0.0&0.0&0.0&0.0&0.0&0.0\\
        \small{Targeted attack} &18.61&13.09&13.55&22.30&15.48&11.99&16.36&20.06&20.47&19.50&21.05&19.86&21.65&20.96&15.23&28.64&15.28&19.40&23.54&16.03&17.73\\
        \hline
    \end{tabular}
    }
    \caption{\textbf{Per-image targeted attack} on artificial ground-truth - This attack is described in \textbf{Per-image targeted attack} paragraph of Section 4.3 in the main paper and the qualitative results are in Figure \ref{figa5:targetedfigure} of the appendix. Our mAP before attack is approximately zero for all targets because we switch the ground truth labels during evaluation. We see an average increase in mAP of around 18 points. This means our adversarial patch successfully switches the detections of quite a few ground truth boxes to the target class. Note that this attack is more challenging than the blindness attack.}
    \vspace*{-0.3cm}
\label{tab3}
\end{table*}

\begin{figure*}[h!]
\setlength{\linewidth}{\textwidth}
\setlength{\hsize}{\textwidth}
  \begin{center}
  \begin{tabular}{ c c c c }

&&&\\

\includegraphics[width=.2\textwidth]{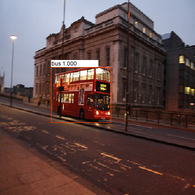}&
\includegraphics[width=.2\textwidth]{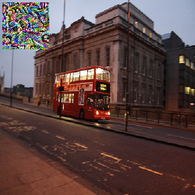}&
\includegraphics[width=.2\textwidth]{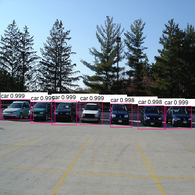}&
\includegraphics[width=.2\textwidth]{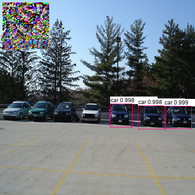}\\

& bus fooled &  & car fooled \\
&&&\\

\includegraphics[width=.2\textwidth]{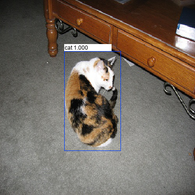}&
\includegraphics[width=.2\textwidth]{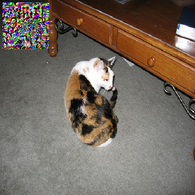}&
\includegraphics[width=.2\textwidth]{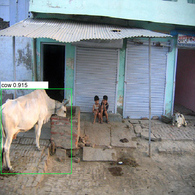}&
\includegraphics[width=.2\textwidth]{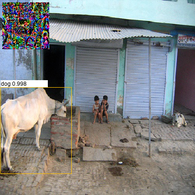}\\

 & cat fooled &  & cow fooled \\
&&&\\

\includegraphics[width=.2\textwidth]{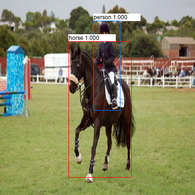}&
\includegraphics[width=.2\textwidth]{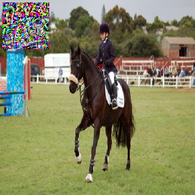}&
\includegraphics[width=.2\textwidth]{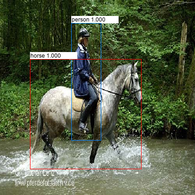}&
\includegraphics[width=.2\textwidth]{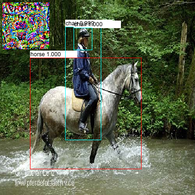}\\

& horse fooled & & person fooled \\
\end{tabular}
  \caption{\textbf{Per-image blindness attack fooling results -} Additional results showcasing the fooling in \textbf{Per-image blindness} attack described in Section 3.1. These are similar to the fooling results in Figure 1 of the main paper. For every pair of columns, the left one is the original image and the right one is the attacked image. The attacked category is written below each example. A failure case is the right image of Row 1, where three out of seven instances of ``cars'' are detected correctly after attack. }
\label{figa2:perimagepatchfigure}
  \end{center}
\end{figure*}
\begin{figure*}[h!]
\setlength{\linewidth}{\textwidth}
\setlength{\hsize}{\textwidth}
  \begin{center}
  \begin{tabular}{ c c c c }

&&&\\

\includegraphics[width=.2\textwidth]{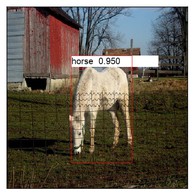}&
\includegraphics[width=.2\textwidth]{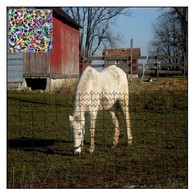}&
\includegraphics[width=.2\textwidth]{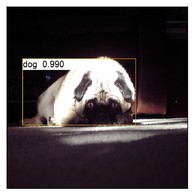}&
\includegraphics[width=.2\textwidth]{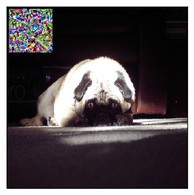}\\

& horse fooled &  & dog fooled \\
&&&\\

\includegraphics[width=.2\textwidth]{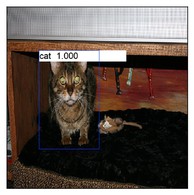}&
\includegraphics[width=.2\textwidth]{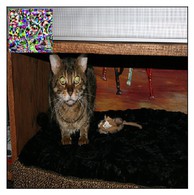}&
\includegraphics[width=.2\textwidth]{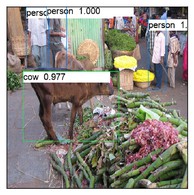}&
\includegraphics[width=.2\textwidth]{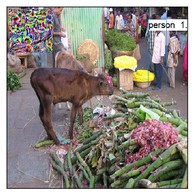}\\

 & cat fooled &  & cow fooled \\
&&&\\

\includegraphics[width=.2\textwidth]{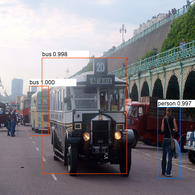}&
\includegraphics[width=.2\textwidth]{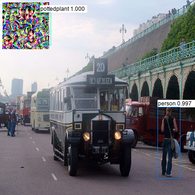}&
\includegraphics[width=.2\textwidth]{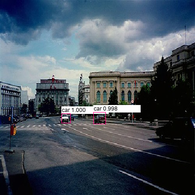}&
\includegraphics[width=.2\textwidth]{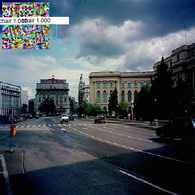}\\

 & bus fooled &  & car fooled \\
&&&\\
\includegraphics[width=.2\textwidth]{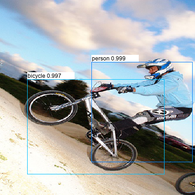}&
\includegraphics[width=.2\textwidth]{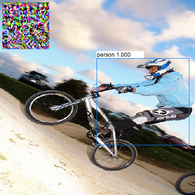}&
\includegraphics[width=.2\textwidth]{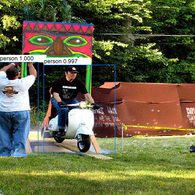}&
\includegraphics[width=.2\textwidth]{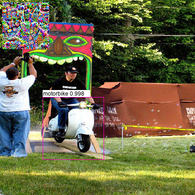}\\
& bicycle fooled &  & person fooled \\
&&&\\
\end{tabular}
  \caption{\textbf{Universal patch blindness attack -} Additional results showcasing \textbf{Universal patch blindness attack} described in Section 3.1 .These are similar to the fooling results in Figure 4 of main paper. For every pair of columns, the left one is the original image and the right one is the attacked image. The patch is always on the top-left corner. The attacked category is written below each example.}
\label{figa3:universalpatchfigure1}
  \end{center}
\end{figure*}
\begin{figure*}[h!]
\setlength{\linewidth}{\textwidth}
\setlength{\hsize}{\textwidth}
  \begin{center}
  \begin{tabular}{ c c c c }

&&&\\
\includegraphics[width=.2\textwidth]{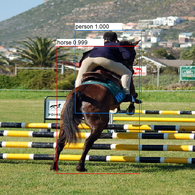}&
\includegraphics[width=.2\textwidth]{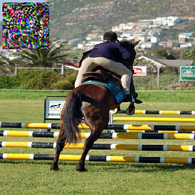}&
\includegraphics[width=.2\textwidth]{{figs/iccv_supplementary/clean/000022}.png}&
\includegraphics[width=.2\textwidth]{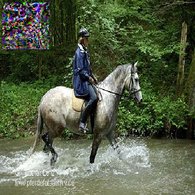}\\

&&&\\

\includegraphics[width=.2\textwidth]{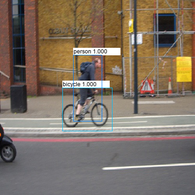}&
\includegraphics[width=.2\textwidth]{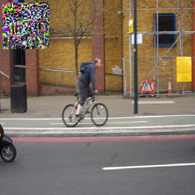}&
\includegraphics[width=.2\textwidth]{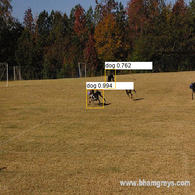}&
\includegraphics[width=.2\textwidth]{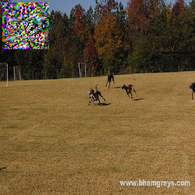}\\

&&&\\

\includegraphics[width=.2\textwidth]{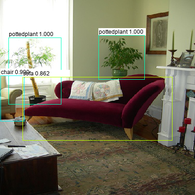}&
\includegraphics[width=.2\textwidth]{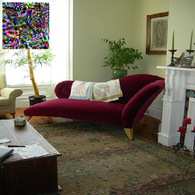}&
\includegraphics[width=.2\textwidth]{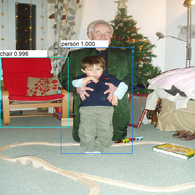}&
\includegraphics[width=.2\textwidth]{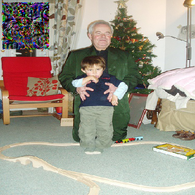}\\

&&&\\

\includegraphics[width=.2\textwidth]{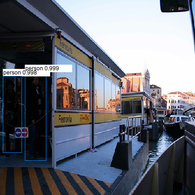}&
\includegraphics[width=.2\textwidth]{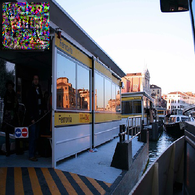}&
\includegraphics[width=.2\textwidth]{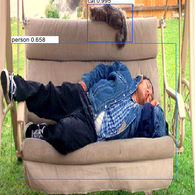}&
\includegraphics[width=.2\textwidth]{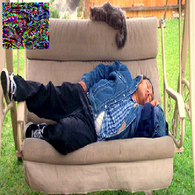}\\

\end{tabular}
  \caption{\textbf{Objectness attack -} This attack is described in \textbf{Per-image objectness attack} paragraph of Section 4.3 in the main paper and the quantitative results are in Table \ref{tab:Objectness attack} of the appendix. For every pair of columns, the left one is the original image and the right one is the attacked image. Note that this attack is class agnostic.} 
\label{figa4:objectnessfigure}
  \end{center}
\end{figure*}

\begin{figure*}[h!]
\setlength{\linewidth}{\textwidth}
\setlength{\hsize}{\textwidth}
  \begin{center}
  \begin{tabular}{ c c c c }

&&&\\

\includegraphics[width=.2\textwidth]{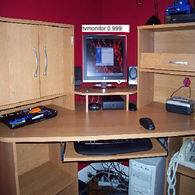}&
\includegraphics[width=.2\textwidth]{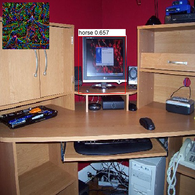}&
\includegraphics[width=.2\textwidth]{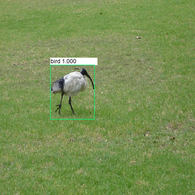}&
\includegraphics[width=.2\textwidth]{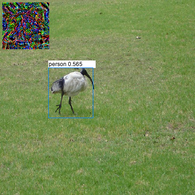}\\

& target: horse &  & target: person \\
&&&\\

\includegraphics[width=.2\textwidth]{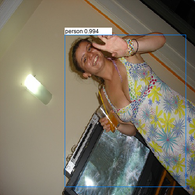}&
\includegraphics[width=.2\textwidth]{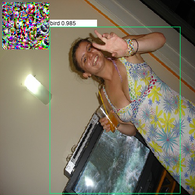}&
\includegraphics[width=.2\textwidth]{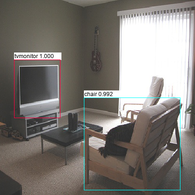}&
\includegraphics[width=.2\textwidth]{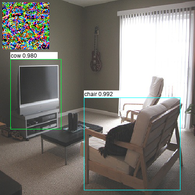}\\

& target: bird &  & target: cow \\
&&&\\

\includegraphics[width=.2\textwidth]{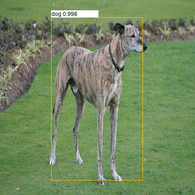}&
\includegraphics[width=.2\textwidth]{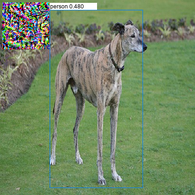}&
\includegraphics[width=.2\textwidth]{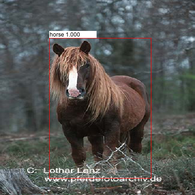}&
\includegraphics[width=.2\textwidth]{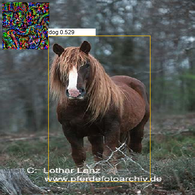}\\
& target: person &  & target: dog \\

\includegraphics[width=.2\textwidth]{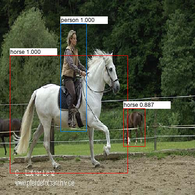}&
\includegraphics[width=.2\textwidth]{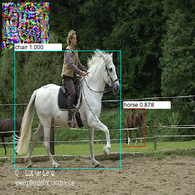}&
\includegraphics[width=.2\textwidth]{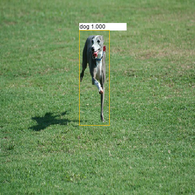}&
\includegraphics[width=.2\textwidth]{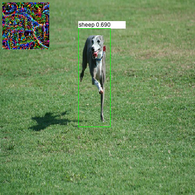}\\
& target: chair &  & target: sheep \\

\end{tabular}
  \caption{\textbf{Per-image targeted attack -} This attack is described in \textbf{Per-image targeted attack} paragraph of Section 4.3 in the main paper and the quantitative results are in Table \ref{tab3} of the appendix. We attack the model to change the label of all objects to the target category. For every pair of columns, the left one is the original image and the right one is the attacked image. The target category is written below each example. As failure cases, a ``horse'' instance on the left image of Row 4 and a ``chair'' instance on the right image of Row 2 are still detected correctly.}
\label{figa5:targetedfigure}
  \end{center}
\end{figure*}

\end{appendices}

\end{document}